\documentclass{article}
\usepackage{iclr2027_conference,times}

\usepackage{amsmath,amsfonts,bm}

\def\eqref#1{equation~\ref{#1}}

\def\1{\bm{1}}

\DeclareMathAlphabet{\mathsfit}{\encodingdefault}{\sfdefault}{m}{sl}
\SetMathAlphabet{\mathsfit}{bold}{\encodingdefault}{\sfdefault}{bx}{n}

\usepackage{hyperref}
\usepackage{url}
\usepackage{amssymb}
\usepackage{xcolor}
\usepackage{amsfonts}
\usepackage{graphicx}
\usepackage{booktabs}
\usepackage{bbm}
\usepackage{enumitem}
\usepackage[T1,OT1]{fontenc}
\usepackage{pifont}
\usepackage{tikz}
\usetikzlibrary{arrows.meta,positioning,shapes.geometric,fit,backgrounds}
\usepackage{amsmath}
\AddToHook{env/table/begin}{\setlength{\belowcaptionskip}{6pt}}
\title{Where RL Breaks: Localizing and Repairing\\ Failures via Targeted Continued Pretraining}

\author{\begin{tabular}{c@{\qquad}c}
  Xinyi Wang & Danqi Chen \\
  \mdseries\texttt{wangxinyilinda@gmail.com} & \mdseries\texttt{danqic@princeton.edu}
\end{tabular} \\
Princeton Language and Intelligence
}

\newcommand{\ngram}[1]{\texttt{#1}}

\iclrpreprintcopy
\begin{document}

\maketitle

\begin{abstract}
Reinforcement learning with verifiable rewards (RLVR) is widely used to improve reasoning in
language models, yet growing evidence suggests that it mostly sharpens what the base model
can already do. This raises a practical question: \emph{when RL fails, what
is the base model missing?} We look inside the RL policy's own rollouts and find that
failures are localized: whether a rollout fails is strongly predicted by a small set of
characteristic $n$-grams, ranging from reflection phrases such as \ngram{let's recheck}
to specific operations such as \ngram{the inverse of A}. These failures persist throughout RL training, although they are not beyond
the model's reach: other sampled answers to the same question are often correct.
Building on this observation, we use the failure $n$-grams to retrieve text from the
pretraining corpus and continue pretraining the base model on it before RL.
The resulting corpus has only 35M tokens, 3--4 orders of magnitude smaller than typical
mid-training. We evaluate this approach on OLMo2 and Qwen3 models trained with RL
on (1) GSM8K; (2) a mixture of math, commonsense, and knowledge questions; and (3) general
math and science reasoning. Despite its size, the corpus yields consistent gains: our
approach outperforms continued pretraining on the same amount of text sampled from the
pretraining corpus across 10 out of 11 settings, including by $2.0$ points on GSM8K when inserted before
mid-training. 
The retrieved text also transfers across model families:
text retrieved for OLMo2-1B's failures improves Qwen3-4B on the mixture by $1.9$ points
over the same baseline. Together, our results suggest that
where RL fails is a cheap and effective signal for selecting pretraining data.
\footnote{Our code is released at \url{https://github.com/WANGXinyiLinda/where-rl-breaks}}
\end{abstract}

\section{Introduction}

Reinforcement learning with verifiable rewards (RLVR) has become the dominant recipe for
improving reasoning in language models. Yet growing evidence suggests that it largely
\emph{sharpens} what the base model can already do: it makes successful solutions more
likely and failing ones less likely, but rarely produces solutions the base model could not
already sample~\citep{yue2026limit, zhao2025echo, wu2025invisible}.
Prolonged or curriculum RL can go further~\citep{liu2025prorl, cai2026curriculum}, but the
base model still sets the limit: RL builds new strategies only from skills the base model
already has~\citep{cheng2025atomic}, and its gains depend on what pretraining
provides~\citep{zhang2025interplay, shen2026pretraining, foster2025foundation}.
This raises a practical question: when RL fails, what exactly is the base model missing,
and can we supply it before RL?

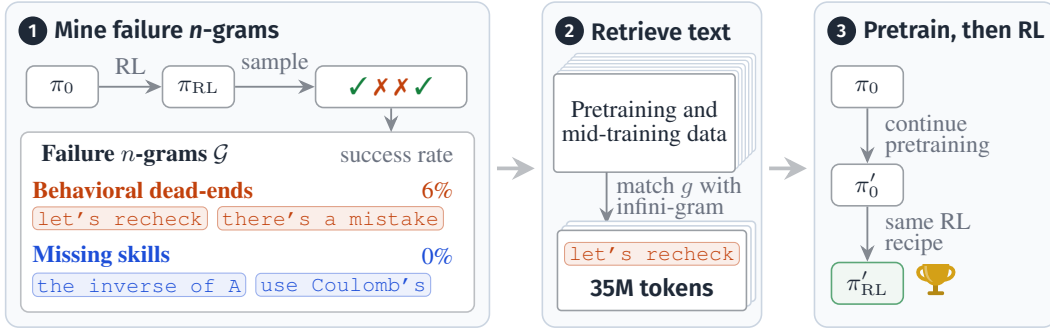
\begin{figure}[t]
    \centering
    \definecolor{fgInk}{HTML}{1F2937}
\definecolor{fgMuted}{HTML}{6B7280}
\definecolor{fgLine}{HTML}{CBD5E1}
\definecolor{fgCard}{HTML}{F8FAFC}
\definecolor{fgBeh}{HTML}{C2410C}
\definecolor{fgSkill}{HTML}{1D4ED8}
\definecolor{fgGood}{HTML}{15803D}
\definecolor{fgGold}{HTML}{D4A017}
\tikzset{trophy/.pic={
  \fill[fgGold!85!black] (-0.07,-0.1) rectangle (0.07,-0.3);
  \fill[fgGold!85!black, rounded corners=0.6pt] (-0.26,-0.3) rectangle (0.26,-0.42);
  \draw[fgGold, line width=1.4pt] (-0.36,0.46) .. controls (-0.66,0.46) and (-0.62,0.08) .. (-0.24,0.06);
  \draw[fgGold, line width=1.4pt] (0.36,0.46) .. controls (0.66,0.46) and (0.62,0.08) .. (0.24,0.06);
  \fill[fgGold] (-0.42,0.56) -- (0.42,0.56) .. controls (0.42,0.02) and (0.2,-0.12) .. (0,-0.12)
                .. controls (-0.2,-0.12) and (-0.42,0.02) .. (-0.42,0.56) -- cycle;
  \fill[white, opacity=0.35] (-0.28,0.5) .. controls (-0.28,0.2) and (-0.18,0.05) .. (-0.08,0.0)
                -- (-0.14,0.0) .. controls (-0.3,0.08) and (-0.36,0.25) .. (-0.36,0.5) -- cycle;
}}
\providecommand{\fgTitleFont}{\fontencoding{T1}\fontfamily{FiraSans-LF}\fontseries{sb}\selectfont}
\begin{tikzpicture}[
  x=1cm, y=1cm, font=\small, text=fgInk,
  >={Stealth[length=2.2mm, width=1.9mm]},
  card/.style={draw=fgLine, fill=fgCard, rounded corners=5pt, line width=0.6pt},
  title/.style={font=\small\fgTitleFont, anchor=west, inner sep=0pt},
  num/.style={circle, fill=fgInk, text=white, font=\scriptsize\fgTitleFont,
              inner sep=0pt, minimum size=3.8mm},
  pi/.style={draw=fgInk!45, fill=white, rounded corners=3pt, line width=0.6pt,
             minimum height=0.55cm, minimum width=0.95cm, inner sep=2pt},
  chip/.style={rounded corners=2pt, inner sep=1.7pt,
               font=\fontsize{8}{9}\selectfont\ttfamily, anchor=west},
  chipB/.style={chip, fill=fgBeh!9, draw=fgBeh!50, text=fgBeh},
  chipS/.style={chip, fill=fgSkill!8, draw=fgSkill!45, text=fgSkill},
  flow/.style={->, line width=0.7pt, draw=fgInk!55},
  lab/.style={font=\small, text=fgMuted},
  big/.style={->, line width=1.4pt, draw=fgInk!30, >={Stealth[length=3mm, width=3mm]}},
]
\draw[card] (0,0)     rectangle (6.4,-4.3);
\draw[card] (7.1,0)   rectangle (10.0,-4.3);
\draw[card] (10.7,0)  rectangle (13.9,-4.3);
\node[num] at (0.36,-0.38) {1};  \node[title] at (0.64,-0.38) {Mine failure \textit{n}-grams};
\node[num] at (7.46,-0.38) {2};  \node[title] at (7.74,-0.38) {Retrieve text};
\node[num] at (11.06,-0.38) {3}; \node[title] at (11.34,-0.38) {Pretrain, then RL};
\draw[big] (6.5,-2.2) -- (7.0,-2.2);
\draw[big] (10.1,-2.2) -- (10.6,-2.2);

\node[pi] (p0) at (0.75,-1.12) {$\pi_0$};
\node[pi] (prl) at (2.55,-1.12) {$\pi_{\mathrm{RL}}$};
\node[pi, minimum width=2.0cm] (roll) at (5.1,-1.12)
     {\textcolor{fgGood}{\ding{51}}\;\textcolor{fgBeh}{\ding{55}}\;\textcolor{fgBeh}{\ding{55}}\;\textcolor{fgGood}{\ding{51}}};
\draw[flow] (p0) -- node[lab, above=1pt] {RL} (prl);
\draw[flow] (prl) -- node[lab, above=1pt] {sample} (roll);

\draw[fill=white, draw=fgInk!30, line width=0.6pt, rounded corners=3pt] (0.2,-1.72) rectangle (6.2,-4.1);
\draw[flow] (roll.south) -- (5.1,-1.72);
\node[anchor=west, font=\small\bfseries] at (0.35,-2.02) {Failure $n$-grams $\mathcal{G}$};
\node[anchor=east, lab] at (6.05,-2.02) {success rate};
\node[anchor=west, font=\small\bfseries, text=fgBeh, inner sep=0pt] (k1) at (0.35,-2.48) {Behavioral dead-ends};
\node[anchor=east, text=fgBeh] at (6.05,-2.48) {6\%};
\node[chipB] (c1) at (0.35,-2.86) {let's recheck};
\node[chipB] at ([xshift=1.2mm]c1.east) {there's a mistake};
\node[anchor=west, font=\small\bfseries, text=fgSkill, inner sep=0pt] (k2) at (0.35,-3.36) {Missing skills};
\node[anchor=east, text=fgSkill] at (6.05,-3.36) {0\%};
\node[chipS] (c3) at (0.35,-3.74) {the inverse of A};
\node[chipS] at ([xshift=1.2mm]c3.east) {use Coulomb's};

\foreach \k in {6,5,...,1}
  \draw[fill=white, draw=fgLine, line width=0.5pt, rounded corners=1.5pt]
       (7.25+0.045*\k,-2.25+0.045*\k) rectangle (9.65+0.045*\k,-0.95+0.045*\k);
\draw[fill=white, draw=fgInk!45, line width=0.6pt, rounded corners=1.5pt]
     (7.25,-2.25) rectangle (9.65,-0.95);
\node[align=center, inner sep=0pt] at (8.45,-1.6) {Pretraining and\\[-1pt]mid-training data};
\draw[flow] (7.95,-2.25) -- node[lab, right, align=left]
      {match $g$ with\\[-2pt]infini-gram} (7.95,-2.94);
\foreach \d in {0.12, 0.06}
  \draw[fill=white, draw=fgLine, line width=0.6pt, rounded corners=1.5pt]
       (7.3+\d,-4.08+\d) rectangle (9.8+\d,-3.02+\d);
\draw[fill=white, draw=fgInk!45, line width=0.6pt, rounded corners=1.5pt]
     (7.3,-4.08) rectangle (9.8,-3.02);
\node[chipB, anchor=center] at (8.55,-3.33) {let's recheck};
\node[font=\small\fgTitleFont, text=fgInk] at (8.55,-3.77) {35M tokens};

\node[pi] (q0) at (11.4,-1.12) {$\pi_0$};
\node[pi] (q1) at (11.4,-2.42) {$\pi_0'$};
\node[pi, draw=fgGood!70, fill=fgGood!6] (q2) at (11.4,-3.72) {$\pi_{\mathrm{RL}}'$};
\pic[scale=0.42] at ([xshift=0.42cm, yshift=0.02cm]q2.east) {trophy};
\draw[flow] (q0) -- node[lab, right=3pt, align=left] {continue\\[-2pt]pretraining} (q1);
\draw[flow] (q1) -- node[lab, right=3pt, align=left] {same RL\\[-2pt]recipe} (q2);
\end{tikzpicture}
    \vspace{-6pt}
    \caption{Overview. We mine failure $n$-grams from RL rollouts, retrieve text containing
    them, continue pretraining on it, and re-run the same RL recipe. The resulting
    $\pi'_{\mathrm{RL}}$ consistently outperforms training on an equal amount of text sampled
    from the same corpora.}
    \label{fig:pipeline}
    \vspace{-12pt}
\end{figure}

Prior work has approached this question from two separate directions. One line studies
where in a reasoning trace RL has its effect or where things go wrong: RL updates
concentrate on a small fraction of high-entropy tokens~\citep{wang2026high}, and a single
token can mark where a reasoning trace turns toward failure~\citep{ko2026cliff}. These
findings have been used to change the training objective~\citep{ko2026cliff}, but not to
choose training data. The other line improves the base model with data before RL. Most
prominently, mid-training on hundreds of billions of reasoning-heavy tokens, now a standard
stage in open training recipes~\citep{olmo2025olmo3}, makes base models far more responsive
to RL~\citep{wang2025octothinker}; continued pretraining on text rich in reasoning
behaviors such as verification and backtracking has a similar effect~\citep{gandhi2025cognitive}.
These corpora, however, are chosen by task similarity or by hand, not by what a particular
model gets wrong. We connect the two directions: like mid-training, we change the data
before RL, but we use a model's RL failures to select a small, targeted corpus.

We approach this question by asking \emph{where} within a rollout RL fails. Our central
observation is that failures are not diffuse: whether a rollout leads to an incorrect outcome is strongly
predicted by whether it contains one of a small number of characteristic $n$-grams (e.g., the
small set of RL-critical tokens of \citealp{wang2026high} and the single-token failure
triggers of \citealp{ko2026cliff}). Which $n$-grams are diagnostic
depends on the model and the task: reflection and self-correction phrases (\ngram{let's
recheck}, \ngram{there's a mistake in the problem}) for weak models on elementary math, where
the model reflects and then spirals rather than recovers; and specific operations (\ngram{convert
the given matrix to reduced row echelon form}, \ngram{the inverse of A}) for stronger models
on broad reasoning suites.
In both cases, the $n$-gram does not cause the failure; it is a visible sign that the
rollout has reached a point from which it rarely recovers. RL has little to learn from at
such points: it improves the model only from rollouts the model itself samples, and
rollouts that reach these points rarely succeed.

These failures are also not beyond the model's reach: for nearly half of the questions
with a failing rollout, another sampled answer to the same question is correct. This
suggests that what the base model lacks is specific, and that a small amount of the right
data could supply it. Our approach follows directly (Figure~\ref{fig:pipeline}): we use the
failure $n$-grams as exact-match queries over the pretraining and mid-training corpora
(Olmo-mix and Dolmino-mix; \citealp{olmo20242}), continue pretraining the base model on the
retrieved text, and re-run the same RL recipe.
We evaluate on OLMo2~\citep{olmo20242} and Qwen3~\citep{yang2025qwen3} models trained with
RL on (1) GSM8K~\citep{cobbe2021gsm8k}; (2) a mixture of math, commonsense, and knowledge
questions (GSM8K, Social IQA, MMLU, and TriviaQA;
\citealp{sap2019social,hendrycks2021measuring,joshi2017triviaqa}); and (3) general math and
science reasoning~\citep{ma2025generalreasoner}, inserting continued pretraining either on
the released base model or, for OLMo2, before mid-training. Our contributions are:
\begin{itemize}[leftmargin=*, topsep=0pt, itemsep=1pt, parsep=0pt]
  \item \textbf{Where RL fails.} RL failures concentrate on a small set of failure
  $n$-grams whose content varies with the model and task, from reflection phrases to
  specific operations. These failures persist throughout RL, even when the model can
  answer the same question correctly (Sections~\ref{sec:observation} and~\ref{sec:local}).
  \item \textbf{A small, targeted fix.} Retrieving pretraining text with the failure
  $n$-grams yields a $35$M-token corpus, 3--4 orders of magnitude smaller than
  mid-training, selected by the model's own failures (Section~\ref{sec:repair}).
  \item \textbf{Consistent gains that transfer.} Our approach outperforms continued
  pretraining on the same amount of text sampled from the same corpus in 10 of 11
  settings, including by $2.0$ points on GSM8K when inserted before mid-training (3 RL
  seeds). A corpus mined from OLMo2-1B's failures improves Qwen3-0.6B and Qwen3-4B by up
  to $2.4$ points.
\end{itemize}

\section{Related Work}
\label{sec:related}

\paragraph{Does RL expand or amplify base-model reasoning?}
RLVR drives progress in reasoning models \citep{guo2025r1, shao2024deepseekmath}, but
whether it expands or merely sharpens the base model is contested. Evidence for sharpening
includes pass@1 gains without pass@$k$ gains \citep{yue2026limit}, convergence toward
pretraining-favored distributions \citep{zhao2025echo}, updates concentrated at a few
high-entropy tokens \citep{wang2026high}, and gains from uninformative rewards when the
base model already has a strategy \citep{shao2025spurious, wu2025contamination}; when rare
correct trajectories vanish, pass@$k$ can even shrink \citep{zhou2026inversion,
yuan2026diversity}. Counter-evidence shows the ceiling is partly a measurement artifact
\citep{wen2025cotpassk} and that prolonged or curriculum RL can move it \citep{liu2025prorl,
cai2026curriculum}. Work on both sides finds that RL composes only skills the base model already has
\citep{cheng2025atomic} and that gains are predicted by pretraining headroom
\citep{shen2026pretraining, zhang2025interplay, foster2025foundation}. Rather than changing
RL, we ask \emph{which} capability the base model lacks and supply it on the pretraining
side.

\paragraph{Mid-training and continued pretraining for RL.}
A growing body of work inserts a mid-training phase between pretraining and RL: hundreds of
billions of math-heavy and chain-of-thought tokens can make Llama as RL-friendly as
Qwen \citep{wang2025octothinker}, open recipes ship such a stage \citep{olmo20242,
olmo2025olmo3}, and controlled studies find mid-training worth more than RL alone
\citep{zhang2025interplay}. Variants place reasoning data in pretraining itself
\citep{thinkaug2025, frontload2025}, run RL during mid-training \citep{rlpt2025,
tian2025rmt}, use an RL-tuned model to shape mid-training data \citep{remit2026,
rrv2026selfgen}, or prime a base model by continued pretraining on verification and
backtracking behaviors \citep{gandhi2025cognitive, shah2025reflection}. We share the
premise that the repair belongs before RL but differ in signal and scale: the model's own
failure $n$-grams select a corpus 3--4 orders of magnitude smaller.
Continual-pretraining studies show that re-warming on a converged model first raises loss
on the original distribution and that replay limits forgetting \citep{gupta2023rewarm,
ibrahim2024simple, abbes2025replay}; Section~\ref{sec:results} shows the same effect.

\paragraph{Pretraining data selection and failure localization.}
Selecting data for a target capability ranges from $n$-gram importance resampling
\citep{xie2023dsir} to gradient- and influence-based methods \citep{xia2024less,
yu2024mates, infreason2025}; complementary work traces behavior to pretraining documents
\citep{taskgram2024, ruis2025procedural}. Our retrieval is a lightweight instance
(exact-match lookup via infini-gram; \citealp{liu2024infinigram}), but the signal is new:
markers of where a trained model breaks down, not similarity to a target task.
On the failure side \citep[see][for a survey]{song2026failures}, intrinsic self-correction is unreliable without external signal
\citep{huang2024selfcorrect, kamoi2024survey}, ``recheck'' steps are overwhelmingly
confirmatory \citep{selfverif2026}, and excess reflection surfaces as overthinking
\citep{sui2025overthinking}, consistent with the reflection phrases among our failure
$n$-grams. Closest to our diagnosis, \citet{ko2026cliff} identify
single cliff tokens at which success probability drops and repair them with preference
optimization; our failure $n$-grams are the state-level, model-specific analogue mined from
RL rollouts, and we use them to retrieve pretraining data rather than shape the RL objective.

\section{Method}
\label{sec:method}

Our method has three parts (Figure~\ref{fig:pipeline}). We first run RL and read off its
rollouts a small set of \emph{failure $n$-grams}, phrases whose presence in a rollout
predicts failure (Section~\ref{sec:localize}). We then argue that RL does little to repair
the states these phrases mark, because an on-policy objective learns how to continue from
them only through the few successful continuations it samples there
(Section~\ref{sec:diagnosis}). Finally, we use the failure $n$-grams as retrieval queries
over the pretraining corpus, continue pretraining on the retrieved text, and re-run the
same RL recipe; a matched control runs the same pipeline with content-neutral queries
(Section~\ref{sec:repair}).

\subsection{Preliminaries}
\label{sec:prelim}

We consider a reasoning task with prompts $x\sim\mathcal{D}$ and a verifier that assigns
each rollout $y$, a reasoning trace followed by an answer, a binary reward
$r(x,y)\in\{0,1\}$. We post-train the base model $\pi_0$ with GRPO
\citep{shao2024deepseekmath}. For each prompt, $G$ rollouts $y_1,\dots,y_G$ are sampled
from the current policy $\pi_{\theta_{\mathrm{old}}}$, and each receives the group-relative
advantage $\hat A_i=\big(r(x,y_i)-\mathrm{mean}_j\,r(x,y_j)\big)/\mathrm{std}_j\,r(x,y_j)$,
which is shared by all of its tokens and is zero when all $G$ rewards are equal. The
policy is updated to maximize
\begin{equation}
  \mathcal{J}(\theta)=\mathbb{E}_{x,\{y_i\}}\Big[\tfrac{1}{G}\textstyle\sum_{i=1}^{G}\tfrac{1}{|y_i|}\sum_{t=1}^{|y_i|}
  \min\big(\rho_{i,t}\hat A_i,\;\mathrm{clip}(\rho_{i,t},1-\epsilon,1+\epsilon)\hat A_i\big)\Big]
  -\beta\,\mathbb{D}_{\mathrm{KL}}\big[\pi_\theta\,\|\,\pi_0\big]
\end{equation}
where $\rho_{i,t}=\pi_\theta(y_{i,t}\mid x,y_{i,<t})/\pi_{\theta_{\mathrm{old}}}(y_{i,t}\mid x,y_{i,<t})$
is the token-level importance ratio, $\epsilon$ the clipping range, and $\beta$ the weight
of a KL penalty toward $\pi_0$ (values, including the asymmetric clipping used for General
Reasoner, are in Appendix~\ref{app:hparams}). We write $\pi_{\mathrm{RL}}$ for the
resulting policy. Because advantages are estimated only from rollouts the policy itself
samples, RL learns how to act in a given state only from the continuations it samples
there; Section~\ref{sec:diagnosis} returns to this point.

\subsection{Localizing failures with $n$-grams}
\label{sec:localize}

We treat $n$-grams as cheap, observable markers of rollout states and ask which ones are
diagnostic of failure. An $n$-gram $g$ is a span of $n$ consecutive tokens, such as
\ngram{let's recheck}, and $E_g(y)=\mathbbm{1}[\,g\text{ appears in }y\,]$ indicates that
rollout $y$ contains it. Over the rollouts of a fixed policy, we measure the
\emph{conditional accuracy} of the rollouts that contain $g$,
\begin{equation}
  a(g)\;=\;\Pr\big[\,r(x,y)=1\;\big|\;E_g(y)=1\,\big].
\end{equation}
We use $\pi_{\mathrm{RL}}$ rather than $\pi_0$, since its rollouts show the failures that
remain after RL. This quantity is associational: whether a rollout emits
\ngram{let's recheck} is itself an outcome of the model's state, so $a(g)$ measures how
strongly $g$ \emph{marks} a low-accuracy state, not the effect of inserting $g$. This
suffices to \emph{localize} failures.

Concretely, we mine $n$-grams from the reasoning spans of $\pi_{\mathrm{RL}}$'s training
rollouts, keep those that occur at least 20 times and in at least $0.1\%$ of the rollouts,
remove candidates dominated by digits or formatting, merge near-duplicates into their
longest common span, and rank the rest by $a(g)$. The lowest-accuracy phrases form the
failure set $\mathcal{G}$, which also serves as the set of retrieval
queries.\footnote{Across settings, $n$ ranges from 3 to 6 and $\mathcal{G}$ contains 4 to
20 queries; Appendix~\ref{app:queries} lists every query verbatim. For the main OLMo2-1B
GSM8K results, we use four representative queries chosen from the top 40 $n$-grams, which
contain many near-duplicates.}

\subsection{Diagnosis: localized low-value states}
\label{sec:diagnosis}

We view a rollout as a trajectory through latent reasoning states: after emitting $t$
tokens, the model is in a state $s_t$ that summarizes the prefix so far. The value of a
state under policy $\pi$,
\begin{equation}
  V^{\pi}(s)=\Pr[\,\text{eventual success}\mid s\,],
\end{equation}
is the probability that a rollout of $\pi$ that passes through $s$ ends with a correct
answer. An $n$-gram connects the two views. We call the states in which the model emits
$g$ the states \emph{marked} by $g$ and write them $s_g$. The rollouts that contain $g$
are exactly those that pass through $s_g$, so the conditional accuracy of
Section~\ref{sec:localize} estimates their average value, $a(g)\approx V^{\pi}(s_g)$, and
failure $n$-grams mark low-value states. These states fall into two archetypes that share
this value structure but differ in what is missing. In an \emph{unproductive behavior},
the model reflects and then spirals rather than recovers: the pieces exist, but their
composition fails. In a \emph{missing skill}, the model stalls on an operation the
subproblem requires, such as row reduction.

RL does little to repair such states. Under GRPO (Section~\ref{sec:prelim}), the policy
learns how to continue from $s_g$ only through the rollouts it samples there, and when
$V^{\pi}(s_g)$ is low, few of them succeed. Section~\ref{sec:local} bears this out: over
RL, the rate at which rollouts enter the failure states stays flat, and the success rate
from them rises only in step with overall accuracy (Figure~\ref{fig:states}), consistent
with the small post-RL shifts reported by \citet{wang2026high} and \citet{zhao2025echo}.
Because on-policy sampling rarely produces the missing signal, we supply it before RL
instead, by continued pretraining on text in which the failure $n$-grams occur. This
changes $\pi_0$ itself, and with it both the policy that RL starts from and the reference
of its KL penalty.

\subsection{Repair: failure-$n$-gram-guided continued pretraining}
\label{sec:repair}

The repair has three steps. (i)~\emph{Retrieve:} each $g\in\mathcal{G}$ is an exact-match
query to an infini-gram index \citep{liu2024infinigram} of the model's pretraining and
mid-training corpora. We extract a 2k-token window around every match, so that the text
includes what surrounds the phrase, and sample documents with a per-query balanced budget
of $35$M tokens in total unless stated otherwise. (ii)~\emph{Continue pretraining:} the
base model is trained on the retrieved text with the next-token objective, either as
released or, for OLMo2, after pretraining stage 1 and before mid-training
(Section~\ref{sec:setup}). (iii)~\emph{Run RL:} GRPO is re-run with all hyperparameters
held fixed, so any difference is attributable to the CPT phase. The \emph{matched control}
runs the identical pipeline and budget with content-neutral queries (e.g.,
\ngram{step 1}), which separates the choice of text from the effect of continued
pretraining itself.

\section{Experiments}
\label{sec:experiments}

We now test the two claims of Section~\ref{sec:method} empirically: that failure $n$-grams
localize where RL breaks, and that continued pretraining on text retrieved with them
repairs it. The experiments span two model families, three RL settings of increasing
breadth, and two insertion points, always against a matched control that receives the
same pipeline and budget with content-neutral queries.
Section~\ref{sec:observation} shows that RL failures concentrate in a small set of failure
$n$-grams; Section~\ref{sec:results} shows that CPT guided by them raises the accuracy RL
reaches over matched controls, most of all before mid-training; Section~\ref{sec:local} shows
that the deficit is conditional rather than a coverage gap; and
Section~\ref{sec:ablations} shows that $35$M tokens suffice.

\subsection{Experimental setup}
\label{sec:setup}

\paragraph{Models and tasks.} We study OLMo2~\citep{olmo20242} at 1B and 7B, as their
pretraining and mid-training corpora are public, and Qwen3~\citep{yang2025qwen3} at 0.6B,
1.7B, and 4B, as they are among the best open-source small models. For OLMo2-1B we also use the public stage-1 checkpoint, taken before
mid-training, which lets us insert the repair \emph{before} mid-training. We always
retrieve from the OLMo2 corpora, Olmo-mix and Dolmino-mix \citep{chen2026olmix},
for all models, including Qwen3, whose own corpus is not public; the Qwen3 runs therefore
also test transfer across pretraining corpora. We consider three RL settings of increasing breadth: (i)
GSM8K~\citep{cobbe2021gsm8k}; (ii) a multi-task mixture of GSM8K, Social
IQA~\citep{sap2019social}, MMLU~\citep{hendrycks2021measuring}, and
TriviaQA~\citep{joshi2017triviaqa}, mixed at a ratio of $8{:}2{:}1{:}1$; and (iii) the
General Reasoner setting~\citep{ma2025generalreasoner}, which trains on
WebInstruct-verified prompts spanning math, science, and logic. All post-training uses
GRPO, with $G=5$ rollouts per prompt for (i) and (ii) (210 and 595 steps) and the General
Reasoner recipe with $G=8$ for (iii) (226 steps). Hyperparameters are held fixed between
the runs with and without CPT.

\paragraph{Continued pretraining and controls.} For each setting we mine failure $n$-grams
from $\pi_{\mathrm{RL}}$ (Section~\ref{sec:localize}), retrieve a corpus with a per-query
balanced budget of $35$M tokens (larger budgets are studied in
Section~\ref{sec:ablations}), and continue pretraining with the next-token objective before re-running RL. We study the effect of inserting CPT before and after OLMo2's mid-training stage. \emph{Before mid-training}: starting
from the OLMo2-1B stage-1 checkpoint, CPT continues the stage-1 optimizer and schedule and
is followed by a shortened stage-2 mid-training (0.84B tokens of the official Dolmino-mix,
identical across conditions). \emph{After mid-training}: CPT is run directly on the released
base model with a fresh optimizer. The \emph{matched control} runs the identical pipeline with the same budget
but retrieves documents with content-neutral random queries. 

\paragraph{Metrics.} We report greedy accuracy at the final RL step, without checkpoint
selection. For GSM8K training we evaluate on the GSM8K test set; for the mixture we report
the macro-average over the GSM8K test set, the MMLU validation set, TriviaQA, and the
Social IQA test set; and for the General Reasoner setting we report the mean over GSM8K,
MATH-500, Minerva, OlympiadBench, AMC, AIME 2024/25, and GPQA-Diamond, graded by
normalized exact match with symbolic equivalence. For the marked states we measure, on
the $G=5$ training rollouts of a step (2560 rollouts), the conditional accuracy $a(g)$,
the entry rate $\Pr[E_g]$, and the fraction of affected questions with a succeeding sibling
rollout; we also report pass@$k$ up to $k=32$. Numbers are single-seed unless marked; where we ran three RL seeds, the standard deviation
across seeds is about one point.

\subsection{Failure \texorpdfstring{$n$}{n}-grams localize RL failures}
\label{sec:observation}

Across all models and settings, a small number of $n$-grams predicts whether a rollout
will fail. For OLMo2-1B on GSM8K, the failure five-grams of Table~\ref{tab:ngrams} appear
in about $7\%$ of late-RL training rollouts, and those rollouts succeed $24\%$ of the time
against $44\%$ for the rest (statistics are always measured on RL steps after the ones the
$n$-grams were mined from, to avoid selection bias). What the failure $n$-grams contain shifts with model strength and task breadth
(Table~\ref{tab:ngrams}). For OLMo2-1B on GSM8K they are reflection and self-correction
phrases (\ngram{there's a mistake}, \ngram{Let's recheck}); for Qwen3-4B on the mixture they are
hedging phrases on the trivia questions (\ngram{a more fitting answer}, \ngram{after careful
consideration}), with which the model talks itself out of an answer; and for the
stronger Qwen3 models in the General Reasoner setting they are specific operations (\ngram{convert the given matrix to
[reduced row echelon form]}, \ngram{use Coulomb's [law]}). The reflection phrases are not rare in pretraining data: each GSM8K retrieval query
matches tens of thousands of Dolmino-mix documents ($57{,}000$ at the median), whereas
the highest-accuracy $n$-grams of the same run (\ngram{pizza is cut into}) match about $950$.
The model has seen \ngram{let's recheck} many times; what it has not learned is what to do
after saying it, so the behavioral deficit is unproductive use rather than absence. Table~\ref{tab:ngrams} also previews the repair: after CPT on text retrieved with these
phrases, the model emits them more often and succeeds more often after them.

\begin{table}[t]
\centering
\small
\caption{Failure $n$-grams mark low-accuracy states, and targeted CPT makes the model
traverse them more often and more successfully. For each phrase we report its conditional
accuracy $a(g)$ and the number of rollouts containing it, late in RL, without and with our
CPT. Summary rows aggregate over the retrieval queries, or over the
failure five-grams mined from the no-CPT run, which were never used for retrieval. General Reasoner runs are compared at matched RL steps.}
\label{tab:ngrams}
\begin{tabular}{llrrrr}
\toprule
& Failure $n$-gram & $a(g)$ & Count & $a(g)$ after CPT & Count after CPT\\
\midrule
\multicolumn{6}{l}{\emph{OLMo2-1B\,/\,GSM8K}} \\
& \ngram{there's a mistake}            & 6.2\%  & 745  & 8.1\%  & 839 \\
& \ngram{Let's recheck}                & 5.6\%  & 591  & 6.3\% & 2166 \\
& \emph{failure five-grams} & 25.7\% & 8859 & 29.2\% & 16620 \\
\midrule
\multicolumn{6}{l}{\emph{Qwen3-4B\,/\,multi-task}} \\
& \ngram{correct answer. The correct}  & 8.7\%  & 126  & 16.7\% & 1360 \\
& \emph{failure five-grams} & 10.8\% & 3123 & 17.6\% & 2140 \\
\midrule
\multicolumn{6}{l}{\emph{Qwen3-1.7B\,/\,General Reasoner}} \\
& \ngram{convert the given matrix to}  & 0.0\% & 232 & 4.3\% & 258 \\
& \ngram{use Coulomb's}                & 0.0\% & 212 & 22.6\% & 997 \\
\multicolumn{6}{l}{\emph{Qwen3-4B\,/\,General Reasoner}} \\
& \ngram{the inverse of A}           & 0.0\% & 56  & 49.9\% & 2219 \\
& \ngram{Then, the derivative of}      & 0.0\% & 44  & 62.1\% & 335 \\
& \emph{failure five-grams} & 18.2\% & 12328 & 36.6\% & 15890 \\
\bottomrule
\end{tabular}
\end{table}
\vspace{-4pt}

\subsection{Targeted continued pretraining repairs RL}
\label{sec:results}

Table~\ref{tab:main} reports the main result. With the RL recipe held fixed, inserting
failure-$n$-gram-guided CPT before RL gives the best final accuracy in every setting of
the table, above both no CPT and the matched control. Over all GSM8K and multi-task
settings we ran, it beats the control in 10 of 11 (9 with 35M tokens and 1 with 100M tokens;
Table~\ref{tab:budget_main}).\footnote{The exceptions are in
Table~\ref{tab:main_rest}.} The gain over the control is largest when
the repair is inserted before mid-training: $+2.0$ on GSM8K over three RL seeds, where the
control falls below no CPT. On released base models the gain is largest for the weakest
model ($+2.4$ over the control for Qwen3-0.6B on GSM8K) and shrinks as the model gets
stronger. 

In the General Reasoner setting, where the failure $n$-grams identify operations rather
than behaviors, the corpus mined from Qwen3-4B and transferred to Qwen3-1.7B leads no CPT
by $+2.6$, and continued pretraining alone, before any RL, raises Qwen3-4B's
eight-benchmark mean by $3.0$ points while the same budget of random Olmo-mix text lowers
it by $4.1$. 

\begin{table}[t]
\centering
\small
\caption{Main results: greedy test accuracy at the final RL step with the RL recipe
fixed. The control uses the identical pipeline and budget with content-neutral queries; all CPT on $35$M tokens. Multi-task is the
macro-average over GSM8K, MMLU, TriviaQA, and Social IQA. General Reasoner is the mean over eight math and science benchmarks. OLMo2-1B GSM8K reports the mean over three RL seeds; others single-seed.
$^\ddagger$:~OLMo2-1B-mined corpus. $^\P$:~compared at a matched RL step (Table~\ref{tab:gr}).}
\label{tab:main}
\begin{tabular}{llccc}
\toprule
& & \multicolumn{3}{c}{Acc after RL}\\
\cmidrule(lr){3-5}
Setting & Model & None & Control & Ours\\
\midrule
\multicolumn{5}{l}{\emph{CPT inserted after stage 1 (before mid-training)}}\\
GSM8K       & OLMo2-1B   & 37.4 & 36.2 & \textbf{38.2} \\
Multi-task  & OLMo2-1B   & 43.2 & 45.3 & \textbf{45.8} \\
\midrule
\multicolumn{5}{l}{\emph{CPT on the released base model (after mid-training)}}\\
GSM8K       & OLMo2-1B   & 59.3 & 60.0 & \textbf{60.6} \\
GSM8K       & Qwen3-0.6B$^\ddagger$ & 74.5 & 73.0 & \textbf{75.4} \\
GSM8K       & Qwen3-4B$^\ddagger$   & 91.9 & 91.7 & \textbf{92.3} \\
Multi-task  & Qwen3-0.6B & 52.4 & 52.2 & \textbf{52.8} \\
Multi-task  & Qwen3-4B$^\ddagger$   & 72.9 & 72.1 & \textbf{74.0} \\
General Reasoner$^\P$ & Qwen3-1.7B & 31.7 & 32.5 & \textbf{34.3} \\
General Reasoner$^\P$ & Qwen3-4B$^\ddagger$ & 44.0 & 43.9 & \textbf{44.7} \\
\bottomrule
\end{tabular}
\end{table}
\vspace{-4pt}

\subsection{The deficit is local, not a global coverage gap}
\label{sec:local}

\paragraph{Failing rollouts share questions with succeeding ones.} Late in RL (OLMo2-1B on GSM8K, steps 150--210), about 70 of the 512
prompts in each training batch have at least one rollout that enters a failure state, and
for $46\%$ of these prompts another rollout of the same question reaches the correct
answer. With only $G=5$ samples per question this is a lower bound, so the correct answer
is within the model's reach for nearly half of the affected questions; what the model
lacks is a way \emph{through} the marked state, not the answer itself.

\begin{figure}[t]
    \centering
    \includegraphics[width=0.9\linewidth]{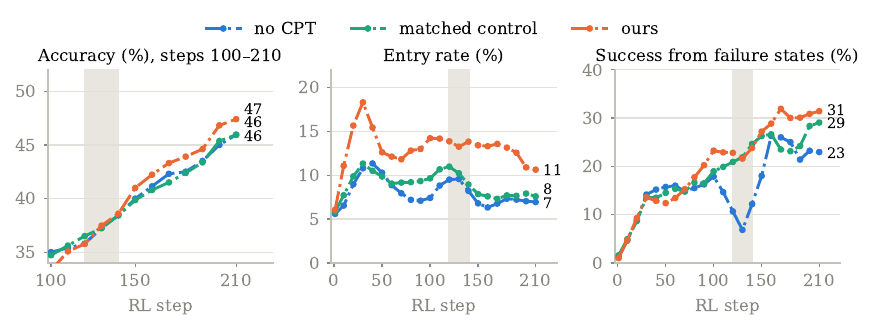}
    \vspace{-15pt}
    \caption{Failure-state statistics over the course of RL for OLMo2-1B on GSM8K with
    CPT inserted after stage 1. The failure set (66 five-grams) is mined at steps 120--140
    (shaded). RL leaves the entry rate flat, whereas our corpus doubles it. The accuracy
    panel starts at step 100, before which the three trajectories coincide; all curves are
    3-point (30-step) moving averages.}
    \label{fig:states}
\end{figure}

\paragraph{RL does not change how often the failure states are entered.}
Figure~\ref{fig:states} tracks the failure states over the course of RL. Training-rollout
accuracy rises from $0.2\%$ to $47\%$ over 210 GRPO steps, but the fraction of rollouts
that enter a failure state stays between $5\%$ and $7\%$ throughout (middle panel), and
the success rate from those states rises at the same pace as overall accuracy (right
panel). RL therefore improves the marked states only in proportion to everything else and does
not single them out for repair \citep{wang2026high, zhao2025echo}. 

\begin{figure}[t]
    \centering
    \includegraphics[width=\linewidth]{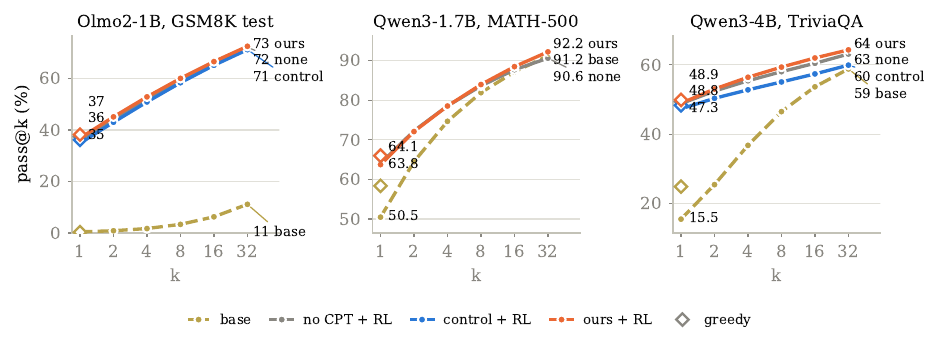}
    \vspace{-20pt}
    \caption{pass@$k$ with 32 samples at $T{=}0.7$; hollow diamonds mark greedy accuracy.
    Left: OLMo2-1B on GSM8K, CPT after stage 1, mean over three RL seeds. Middle:
    Qwen3-1.7B on MATH-500 in the General Reasoner setting (step 226; ours is the
    self-mined corpus, which has no matched control). Right: Qwen3-4B on TriviaQA after
    multi-task RL. The OLMo2-1B base rarely produces the required \texttt{\#\#\#\#}
    answer, hence its near-zero curve. RL raises pass@1 far more than pass@32; the repaired
    model leads the control at every $k$ and, relative to no CPT, leads on OLMo2-1B and
    Qwen3-4B and, at $k{=}32$, edges above it on Qwen3-1.7B.}
    \label{fig:passk}
\end{figure}

\paragraph{RL does not expand coverage; CPT does, modestly.} Figure~\ref{fig:passk} plots
pass@$k$, the fraction of test questions solved by at least one of $k$ samples. Where the
base model already produces the answer format (the middle and right panels), RL raises
pass@1 by a large margin but pass@32 by only a few points, and on MATH-500 pass@32 after
RL ends slightly below the base model's ($90.6$ vs.\ $91.2$), reproducing the finding of
\citet{yue2026limit} that RL makes the model sample its existing solutions more reliably
rather than reach new ones. Continued pretraining before RL moves this ceiling, by a
small amount. The repaired model has higher pass@$k$ than the matched control at every
$k$, and its pass@32 ends above the no-CPT run by $1.2$ points on TriviaQA and $1.6$ on
MATH-500, where it is the only RL-trained model to finish above the base model ($92.2$). On TriviaQA its lead over the control grows with $k$, from $1.6$ points at $k{=}1$ to $4.4$
at $k{=}32$, while its lead over no CPT stays near one point. In short, generic CPT shrinks the set of answers RL can reach, RL itself does not enlarge
it, and targeted CPT enlarges it by one to two points at $k{=}32$, at the edge of what
500 to 1000 test questions can resolve.

\paragraph{Targeted CPT makes the model enter the marked states more often and succeed
from them more often.} Inserting our CPT after stage 1 and re-running the same RL recipe
moves all three quantities of Table~\ref{tab:states_main} in the intended direction. The
entry rate into the failure states doubles, the success rate from them rises by about
four points, and the share of affected questions with a correct sibling rollout rises
from $46\%$ to $50\%$ (Figure~\ref{fig:states}, Table~\ref{tab:states_main}). The success
rate rises in every other setting of Table~\ref{tab:states_main} as well, most strongly
for OLMo2-7B on the mixture. Two observations qualify this result: the matched control raises the success rate by a
similar amount, though without raising the entry rate, and per phrase the reflection
markers succeed $5$--$12\%$ of the time with or without the repair
(Table~\ref{tab:ngrams}; all 16 GSM8K queries in Table~\ref{tab:queries_ag}). What the repair changes most is therefore how often the model
emits these phrases, so its accuracy gain cannot be attributed to the diagnosed states
alone.

\paragraph{Missing skills are repaired more fully than unproductive behaviors.} The two
archetypes of Section~\ref{sec:diagnosis} respond differently to the same repair
(Table~\ref{tab:ngrams}). The operations that Qwen3-4B never completed correctly go from
$0\%$ to $50$--$62\%$ success after CPT, whereas the reflection phrases of OLMo2-1B move
by only one or two points. A missing operation has a definite correct execution, which
the retrieved text can supply directly, whereas a reflection phrase helps only if the
model also knows what went wrong on the question at hand, which the retrieved recoveries
do not teach. The repair therefore raises how often the model
reflects far more than how often reflecting succeeds.

\vspace{-6pt}

\begin{table}[t]
\centering
\small
\caption{Entry rate into the failure states, success rate from them, and test accuracy
before $\to$ after CPT on $35$M retrieved tokens, same RL recipe. After stage 1 and for
OLMo2-7B all three rise; in the General Reasoner setting the success rate rises at an
unchanged entry rate. }
\label{tab:states_main}
\begin{tabular}{llrrr}
\toprule
Setting  & Model & $\Pr[E_{\mathcal{G}}]$ & $a(\mathcal{G})$ & test acc.\ \\
\midrule
GSM8K & OLMo2-1B (after stage 1) & $5.7\to10.6$ & $25.7\to29.2$ & $\mathbf{36.2\to39.2}$ \\
GSM8K & OLMo2-1B (after stage 1, 3 seeds) & $6.8\to7.6$ & $24.3\to28.4$ & $\mathbf{37.4\to38.2}$ \\
General Reasoner & Qwen3-1.7B (base) & $7.4\to7.3$ & $\mathbf{16.9\to27.7}$ & $34.1\to35.0$ \\
\bottomrule
\end{tabular}
\end{table}
\vspace{-4pt}

\subsection{Analysis and ablations}
\label{sec:ablations}

\paragraph{Insert the repair before mid-training.} Table~\ref{tab:targeting} compares the
two insertion points at equal budget and with the identical pipeline. Inserted after
stage 1, the targeted corpus is the best CPT data in both settings: on GSM8K it leads the
matched control by $+2.0$ over three seeds, about two standard deviations, while generic
text at the same budget falls below no CPT; on the mixture it leads no CPT by $+2.6$.
Inserted into the released base model, the same corpus adds $1.3$ points on GSM8K, but
the matched control adds nearly as much, and on the mixture both CPT variants fall below
no CPT. Two effects favor the earlier insertion: targeting only pays off before
mid-training, and the mid-training that follows the repair absorbs the perturbation that
CPT on a finished model leaves in place.

\begin{table}[t]
\centering
\small
\caption{Where to insert the repair: greedy accuracy of OLMo2-1B at the final RL step with
the same $35$M-token pipeline at both insertion points; the bottom rows give the gain of
our corpus over no CPT and over the control. }
\label{tab:targeting}
\begin{tabular}{lcccc}
\toprule
 & \multicolumn{2}{c}{\begin{tabular}{@{}c@{}}After stage 1\\(before mid-training)\end{tabular}} & \multicolumn{2}{c}{\begin{tabular}{@{}c@{}}Released base\\(after mid-training)\end{tabular}} \\
\cmidrule(lr){2-3}\cmidrule(lr){4-5}
CPT data & GSM8K & multi-task & GSM8K & multi-task \\
\midrule
none                             & 37.4 & 43.2 & 59.3 & \textbf{52.6} \\
matched control                  & 36.2 & 45.3 & 60.0 & 49.5 \\
failure $n$-grams (ours)         & \textbf{38.2} & \textbf{45.8} & \textbf{60.6} & 51.5 \\
\midrule
ours $-$ none                    & $+0.8$ & $+2.6$ & $+1.3$ & $-1.1$ \\
ours $-$ matched control         & $+2.0$ & $+0.5$ & $+0.6$ & $+2.0$ \\
\bottomrule
\end{tabular}
\vspace{-6pt}
\end{table}
\vspace{-4pt}

\paragraph{A small corpus is enough.} Table~\ref{tab:budget_main} varies the token budget.
In every setting the gain appears at $35$--$100$M tokens, and scaling the corpus to $1$B
leaves accuracy flat or lowers it. The cost of a large budget is sharpest before RL,
where $1$B tokens of retrieved text lowers Qwen3-4B's eight-benchmark mean by $19$
points; mixing random documents into a large corpus does not recover the loss
(Table~\ref{tab:replay}).

\begin{table}[t]
\centering
\small
\caption{CPT token budget: greedy accuracy at the final RL step against the number of
retrieved tokens; the last row evaluates the CPT model before RL. OLMo2-1B GSM8K at $0$ and $35$M:
three-seed means. }
\label{tab:budget_main}
\begin{tabular}{lllcccc}
\toprule
Task & Model & Insertion & 0 & 35M & 100M & 1B \\
\midrule
GSM8K & OLMo2-1B        & After stage 1 & 37.4 & \textbf{38.2} & 37.7 & 33.4 \\
GSM8K & OLMo2-1B        & Released base & 59.3 & \textbf{60.6} & 60.5 & 59.9 \\
Multi-task & Qwen3-0.6B & Released base & 52.4 & \textbf{52.8} & 52.6 & 51.7 \\
Multi-task & Qwen3-1.7B & Released base & 62.4 & 62.5 & \textbf{63.4} & 63.2 \\
General Reasoner & Qwen3-4B & Released base & 37.5 & \textbf{40.5} & 35.2 & 18.8 \\
\bottomrule
\end{tabular}
\vspace{-6pt}
\end{table}
\vspace{-8pt}

\paragraph{The corpus transfers across model families.} The $35$M corpus mined from
OLMo2-1B's GSM8K failures improves Qwen3-0.6B on GSM8K by $+2.4$ over the control and
Qwen3-4B on the mixture by $+1.9$ (Table~\ref{tab:main}), and in the other direction the
corpus mined from Qwen3-4B's General Reasoner failures lifts Qwen3-1.7B by $+2.6$ over
no CPT (Table~\ref{tab:main}).

\section{Conclusion}
\label{sec:conclusion}

In this paper, we asked what specific capability a base model is missing when RL almost works, and whether
it can be supplied before RL. Failure $n$-grams read off the model's own rollouts mark
low-value states that on-policy, KL-anchored RL cannot self-repair. Using those $n$-grams
to retrieve and continue-pretrain on text that traverses the marked states improves
downstream accuracy over matched controls when inserted before mid-training, and the corpus
transfers across model families, all without labels or a teacher. A model's own failures
are an underused supervision signal: a cheap, label-free pointer to the pretraining data it
needs most.

\bibliography{iclr2027_conference}
\bibliographystyle{iclr2027_conference}

\newpage
\appendix
\section{Limitations}
\label{sec:limitations}

Several limitations scope our claims.

\paragraph{Effect sizes and evaluation.} Gains, while consistent in direction against no
CPT across the GSM8K settings and most multi-task settings (and against the matched
control in all but one), are modest in absolute terms ($0.1$--$2.6$ points);
only the OLMo2-1B GSM8K conditions are averaged over three RL seeds, and some benchmarks
(the small competition-math sets used in the General Reasoner setting) are high-variance.
The seed standard deviation is about one point on GSM8K for a 1B model
(Appendix~\ref{app:seeds}), so no single comparison exceeds two standard deviations, and
the single-seed gains for the stronger Qwen3 models are not individually significant.
In the General Reasoner setting, runs are compared at matched RL steps rather than at
convergence. Stronger conclusions require more seeds and completed
runs.

\paragraph{Dependence on open training corpora.} Our pipeline retrieves repair data from
the base model's own pretraining and mid-training corpora, which restricts it to model
families that release this data (here, OLMo2). For Qwen3 we retrieve from Olmo's corpora
instead, which worked in our experiments but conflates corpus transfer with the repair
itself. Applying the method to models with undisclosed corpora would in general require a
substitute source of correct-traversal text---a public corpus indexed in place of the
original, or synthetically generated traversals---which we do not study here.

\paragraph{Generic in-domain data is a strong baseline.} On an already mid-trained model,
$35$M tokens of a curated math corpus (MegaMath-Web-Pro) matched our targeted corpus on
GSM8K (Table~\ref{tab:targeting_rest}). Targeting clearly beats \emph{matched} generic data and is label-free, but when a
high-quality in-domain corpus already exists, it is the cheaper option; our advantage is
clearest before mid-training and for deficits with no obvious curated corpus (e.g.,
recovery from a self-identified error).

\paragraph{Simplicity of the retrieval signal.} We use exact-match $n$-gram lookup via
infini-gram, a deliberately lightweight instrument. It will miss relevant text that
expresses the same marked state in different surface form (paraphrases, alternative
notation, or different variable names), it can over-fragment a single long pattern into
several near-duplicate $n$-grams, and on task mixtures the lowest-accuracy $n$-grams can be
dominated by artifacts of individual sub-tasks (formatting fragments, rare tokens) rather
than reasoning states, which we currently remove by hand. Retrieved spans also contain
the trigger phrase more reliably than a completed recovery (Section~\ref{sec:local}).
Denser or semantic retrieval seeded by the failure $n$-grams, and filtering retrieved
spans for the presence of a resolution, are natural extensions.

\paragraph{Associational, not causal, diagnosis.} The failure $n$-grams \emph{mark}
low-value states rather than cause them: whether a rollout emits a given phrase is itself
an outcome of the model's state on that sample. We mitigate the most obvious confound by
comparing rollouts of the same question with and without the $n$-gram, but a fuller causal
account---e.g., intervening to inject a marked phrase and measuring downstream
success---remains open, and our claims should be read as identifying \emph{where} failures
concentrate rather than attributing causation to surface tokens.

\paragraph{Scope of the studied regime.} We focus on the regime where RL almost
works---where the correct answer is often within the base model's global support and the
deficit is conditional recovery. Whether the same diagnose-and-repair recipe helps when the
base model is far from competent (a genuine global coverage gap), how it composes over
repeated diagnose--repair--RL cycles, and why large CPT budgets hurt when replay of the
pretraining distribution does not repair them (Section~\ref{sec:ablations}), are questions
we leave to future work.

\section{Hyperparameters}
\label{app:hparams}

\begin{table}[h]
\centering
\small
\caption{RL and continued-pretraining hyperparameters. GRPO uses the low-variance KL
estimator; ``clip-higher'' denotes the asymmetric clipping of DAPO with
$\epsilon_{\text{high}}=0.3$. CPT with the ``after stage 1'' insertion uses the OLMo
trainer and continues the stage-1 schedule; all other CPT uses a fresh AdamW optimizer.}
\label{tab:hparams}
\begin{tabular}{lccc}
\toprule
 & GSM8K & multi-task & General Reasoner \\
\midrule
rollouts per prompt $G$ & 5 & 5 & 8 \\
prompts per step & 512 & 512 & 1024 \\
mini-batch & 128 & 128 & 256 \\
learning rate & $10^{-6}$ & $10^{-6}$ & $5\times10^{-7}$ \\
clip $\epsilon$ & 0.2 & 0.2 & 0.3 (clip-higher) \\
KL weight $\beta$ & $10^{-3}$ & $10^{-3}$ & $10^{-4}$ \\
entropy bonus & 0 & 0 & $10^{-3}$ \\
max prompt / response tokens & 1024 / 1024 & 1024 / 1024 & 1024 / 2048 \\
sampling temperature & 1.0 & 1.0 & 1.0 \\
RL steps (epochs) & 210 (15) & 595 (1) & 226 (1) \\
\midrule
CPT learning rate (released base) & \multicolumn{3}{c}{$5\times10^{-5}$ peak, 10\% warmup, cosine decay, AdamW ($\beta_2{=}0.95$, wd $0.01$)} \\
CPT learning rate (after stage 1) & \multicolumn{3}{c}{$7.4\times10^{-5}$, continuing the stage-1 linear schedule} \\
CPT sequence length / batch & \multicolumn{3}{c}{4096 tokens / 512 sequences (${\approx}2.1$M tokens per step)} \\
CPT steps at 35M / 100M / 1B tokens & \multicolumn{3}{c}{16 / 47 / 476} \\
shortened stage 2 (after stage 1 insertion) & \multicolumn{3}{c}{400 steps $\times$ 512 $\times$ 4096 ${\approx}0.84$B Dolmino tokens} \\
\bottomrule
\end{tabular}
\end{table}

\paragraph{Mining and retrieval.} We mine $n$-grams ($n=3$--$6$) from the reasoning span
of GRPO training rollouts at the final step (2560 rollouts per step, last 50 steps for the state-level statistics), keep those that occur at least 20 times and in at least $0.1\%$ of the rollouts, remove candidates dominated by digits or formatting, merge near-duplicates
(Jaccard $\geq0.9$) into their longest common span, and rank by accuracy. The
lowest-accuracy 10--40 spans are used verbatim as exact-match queries to a local
infini-gram index of Dolmino-mix and OLMoE-mix (DCLM and non-DCLM shards); each match is
extracted with a 2000-token window, queries with fewer than 1000 matching documents are
dropped, and documents are sampled with an equal per-query token budget (up-sampling a
query's pool at most once when it is exhausted).

\section{Retrieval queries behind every corpus}
\label{app:queries}

Table~\ref{tab:queries} lists, verbatim, the exact-match queries used to build each CPT corpus and control. Queries are the merged common spans of Section~\ref{sec:localize}; truncated word pieces (e.g., \ngram{stitute back}) are kept as mined.

\begin{table}[h]
\centering
\scriptsize
\sloppy
\caption{Retrieval queries. ``API index'' denotes the public infini-gram \texttt{v4\_olmo-mix-1124\_llama} index; the others are local indexes of the named corpora.}
\label{tab:queries}
\begin{tabular}{p{0.30\linewidth}p{0.11\linewidth}p{0.53\linewidth}}
\toprule
Corpus & Index & Queries \\
\midrule
GSM8K, OLMo2-1B, after stage 1 (ours, 35M; also the Qwen3 GSM8K and multi-task ``OLMo2-1B-mined'' corpora) & Olmo-mix (API index) & ``let's correct'', ``still incorrect'', ``solve it correctly'', ``wait'' \\
Matched control for the above (35M; also the Qwen3 controls) & Olmo-mix (API index) & ``1'', ``first'', ``step 1'' \\
GSM8K, OLMo2-1B, after stage 1, 175M (ours: 20 lowest-accuracy 6-grams) & Olmo-mix & ``, which totals to '', ``seems there's a mistake'', ``'s a mistake in the'', ``there's a mistake in'', ``, which is not possible.'', ``.\textbackslash\{\}nWe are given that'', ``the nearest whole number, which'', ``recheck the steps.\textbackslash\{\}n'', ``Let's recheck the'', ``. Let's recheck'', ``.\textbackslash\{\}nHe buys '', ``be a whole number, and'', ``must be a whole number,'', ``whole number, which is '', ``we can't have a'', ``can't have a fraction'', ``t have a fraction of a'', ``ving for x, we get'', ``there was a misunderstanding in'', ``standing in the problem statement.'' \\
GSM8K, OLMo2-1B, after stage 1, 175M control (20 highest-accuracy 6-grams) & Olmo-mix & ``, he needs to buy '', ``as follows:\textbackslash\{\}nLet''', ``months in a year, so'', ``find out how many packs'', ``. So, to find out'', ``lollipops.\textbackslash\{\}n'', ``miles.\textbackslash\{\}nTo find out'', ``"\textbackslash\{\}nThe original price of'', ``they eat a total of '', ``ounces.\textbackslash\{\}n\#\#\#\#'', ``.\textbackslash\{\}nHe gets paid \$'', ``.\textbackslash\{\}nNow, let''', ``, we need to subtract the'', ``uesday.\textbackslash\{\}n\#\#\#\# '', ``out how many packs he'', ``pieces. So, the total'', ``.\textbackslash\{\}nHe sells each'', ``.\textbackslash\{\}nSo, the cost'', ``cubic inches.\textbackslash\{\}n\#\#\#\#'', `` so he ate '' \\
GSM8K, OLMo2-1B released base (self-mined, $n{=}5$) & Dolmino-mix, OLMoE-mix & ``which is incorrect'', ``Let's correct'', ``there's a mistake'', ``doesn't make sense'', ``However, let's'', ``Let's recheck'', ``have a negative number of'', ``we'll round'', ``can't have half'', ``we know it takes'', ``or a mistake in'', ``because we can't'', ``there's a misunderstanding'', ``However, based on the'', ``need to recheck the'', ``be a mistake in'' \\
Multi-task, OLMo2-1B released base ($n{=}4$) & Dolmino-mix, OLMoE-mix & ``to calculate the area'', ``a negative number of'', ``Let's re'', ``recheck'', ``: we are given'', ``in the problem statement'', ``the nearest whole number'', ``recheck the steps'', ``since we can'', ``a mistake in the'', ``it seems there'', ``a fraction of a'', ``we set up'', ``find out how long'', ``set up the equation'', ``'t make sense'' \\
Multi-task, Qwen3-0.6B ($n{=}4$) & Dolmino-mix, OLMoE-mix & ``we don't'', ``a fraction of'', ``the nearest whole number'', ``recheck the'', ``doesn't make sense'', ``to find the value'', ``total distance for the'', ``the week is the'', ``Set up the'', ``Elvis Presley'', ``Tiger Woods'', ``Neil Armstrong'', ``The Simpsons'', ``Dwight D. Eisenhower'', ``Jeremy Corbyn'', ``Great Expectations'', ``Marilyn Monroe'', ``Neville Chamberlain'', ``Gone with the Wind'', ``Vatican City'' \\
Multi-task, Qwen3-1.7B ($n{=}4$) & Dolmino-mix, OLMoE-mix & ``Thus, the correct'', ``is described as'', ``life. Therefore,'', ``ilerden hangisi'', ``This option suggests that'', ``unreliable'', ``author's surprise'', ``was not fully aware of'', ``didReceiveMemory'', ``The passage describes'', ``let's determine the'', ``that won the'', ``process is that'', ``in New York City,'', ``that completes the'', ``is that the black'', ``new words is'' \\
General Reasoner, Qwen3-4B ($n{=}3$) & Dolmino-mix, OLMoE-mix & ``stitute back'', ``\& \textbackslash\{\}frac'', ``+ \textbackslash\{\}sqrt'', ``matrix, we'', ``\^\{\}T A'', ``a matrix,'', ``is the original'', ``A\^\{\}T'', ``= \textbackslash\{\}cos'' \\
General Reasoner, Qwen3-4B ($n{=}5$) & Dolmino-mix, OLMoE-mix & ``Then, we'', ``of \^\{\}T'', ``inverse of '', ``Next, we find'', ``Substitute back'', ``given matrix to reduced row'', ``echelon form,'', ``the given matrix to reduced'', ``invert the given matrix to'', ``product of '', ``Then, the derivative of'', ``T\$ and '', ``we find the inverse'', ``the given matrix is:'' \\
General Reasoner, Qwen3-1.7B ($n{=}3$) & Dolmino-mix, OLMoE-mix & ``the given matrix'', ``\textbackslash\{\}Delta T ='', ``Coulomb'', ``\& \textbackslash\{\}frac'', ``\$ is the'', ``matrix, we'', ``The general'', ``\textbackslash\{\}int'', ``the distance between'', ``'s constant'', ``pi r\^\{\}'', ``change in temperature'', ``the magnitude'' \\
General Reasoner, Qwen3-4B, validation-mined ($n{=}5$; the 35M/100M/1B budget runs) & Dolmino-mix & ``Given the'', ``Therefore, the value'', ``solve the problem, we'', ``Human:'', ``is given by:'', ``Substitut'', ``To find the'', ``, we find that the'', ``.\textbackslash\{\}n\textbackslash\{\}nHowever,'', ``re-examine'', ``However, let's'', ``Therefore, there'', ``Let's re-'', ``So, the'', ``didn't want to'', ``the question asks for the'' \\
\bottomrule
\end{tabular}
\end{table}

\section{Run-to-run variation}
\label{app:seeds}

For OLMo2-1B on GSM8K we ran three RL seeds (different data order; vLLM sampling is not
seeded) per condition with the CPT checkpoint fixed. Final-step accuracies are
$36.2/38.2/37.9$ (stage 1, no CPT), $35.6/35.8/37.1$ (matched control), $39.2/38.9/36.5$
(ours); $58.5/59.9/59.6$ (released base, no CPT), $58.9/60.8/60.3$ (random Olmo-mix),
$58.1/60.1/60.0$ (high-accuracy $n$-grams), $60.7/59.6/61.5$ (ours); and, over two seeds,
$61.4/59.8$ (trigger-only) and $59.9/60.3$ (windows with a resolution). The standard
deviation within a condition is $0.7$--$1.5$ points, and the final-checkpoint accuracy of
a single run fluctuates by $1$--$2.5$ points between adjacent evaluation steps (e.g.,
$58.5$ at step 210 vs.\ $61.0$ at step 170). We therefore treat differences below about
one point as null; the stage-1 gain over the matched control ($+2.0$) is about two
standard deviations; the gain over no CPT ($+0.8$) is within noise. All other numbers in
the paper are single-seed, except the OLMo2-1B multi-task no-CPT and ours entries
($52.9/52.3$ and $51.4/51.6$ over two seeds; the mixture's Social IQA component moves by
three to five points between adjacent checkpoints, so single runs there carry about
two points of noise).

\section{Additional results}
\label{app:more}

\begin{table}[h]
\centering
\small
\caption{Multi-task settings in which CPT did not help, omitted from Table~\ref{tab:main}
(same protocol: greedy macro-average at the final RL step; OLMo2-1B
no-CPT and ours: mean $\pm$ s.d.\ over two RL seeds; rest single-seed).}
\label{tab:main_rest}
\begin{tabular}{llccc}
\toprule
Setting & Model & Base$+$RL & Base$+$control CPT$+$RL & Base$+$our CPT$+$RL\\
\midrule
multi-task  & OLMo2-1B   & \textbf{52.6$\pm$0.4} & 49.5 & 51.5$\pm$0.1 \\
multi-task  & Qwen3-1.7B & 62.4 & \textbf{63.4} & 62.5 \\
\bottomrule
\end{tabular}
\end{table}

\begin{table}[h]
\centering
\small
\caption{General Reasoner setting: mean greedy accuracy over eight benchmarks (GSM8K,
MATH-500, Minerva, OlympiadBench, AMC, AIME 2024, AIME 2025, GPQA-Diamond). RL runs with
self-mined corpora were stopped by compute limits before 226 steps, so checkpoints are
compared at the indicated step; the three Qwen3-1.7B runs were later completed to 226
steps. Answers are graded by normalized exact match with symbolic equivalence
(math-verify). AIME and AMC contain 30--40 problems each, so differences below ${\sim}2$
points are noise: at step 226 the three Qwen3-1.7B runs are within $1$ point on GSM8K
($80.1$--$81.2$) and MATH-500 ($66.0$--$66.6$), and their means differ through AMC and GPQA.}
\label{tab:gr}
\resizebox{0.9\linewidth}{!}{%
\begin{tabular}{llcc}
\toprule
Model & CPT corpus & RL step & mean acc.\ \\
\midrule
Qwen3-4B   & none                                          & 35  & 44.0 \\
Qwen3-4B   & reflection queries (OLMo2-1B-mined), 35M      & 35  & 44.7 \\
Qwen3-4B   & self-mined $n{=}3$, 35M                        & 60  & 43.4 \\
Qwen3-4B   & self-mined $n{=}5$, 35M                        & 60  & 43.1 \\
Qwen3-4B   & self-mined $n{=}3{+}5$, 35M                    & 40  & 44.5 \\
Qwen3-4B   & self-mined $n{=}3$, 100M                       & 40  & 42.2 \\
Qwen3-4B   & self-mined $n{=}3$, 1B                         & 45  & 34.2 \\
\midrule
Qwen3-1.7B & none                                          & 100 & 34.1 \\
Qwen3-1.7B & none                                          & 226 & 31.7 \\
Qwen3-1.7B & self-mined $n{=}3$, 35M                        & 100 & 31.0 \\
Qwen3-1.7B & self-mined $n{=}3$, 35M                        & 226 & 32.9 \\
Qwen3-1.7B & Qwen3-4B-mined $n{=}3$, 35M (transfer)         & 100 & 35.0 \\
Qwen3-1.7B & Qwen3-4B-mined $n{=}3$, 35M (transfer)         & 226 & \textbf{34.3} \\
Qwen3-1.7B & Qwen3-4B-mined $n{=}5$, 35M (transfer)         & 95  & 31.9 \\
Qwen3-1.7B & Qwen3-4B-mined $n{=}3$, 100M (transfer)        & 60  & 33.8 \\
Qwen3-1.7B & Qwen3-4B-mined $n{=}3$, 1B (transfer)          & 70  & 34.8 \\
\bottomrule
\end{tabular}}
\end{table}

\begin{table}[h]
\centering
\small
\caption{Qwen3-4B and Qwen3-1.7B before RL (the source of the before-RL rows of Tables~\ref{tab:main} and~\ref{tab:budget_main}): greedy accuracy of the base model and of the CPT models,
same eight-benchmark suite and grader as Table~\ref{tab:gr}. The transfer row is the
OLMo2-1B-mined corpus of Table~\ref{tab:main}; targeted corpora are
self-mined from Dolmino-mix ($n{=}3$: the corpora of Table~\ref{tab:gr}; $n{=}5$: a set mined from
validation rollouts); random = uniformly sampled Olmo-mix
documents; mix = targeted and random documents shuffled together. The lower block
repeats the sweep for Qwen3-1.7B-Base.}
\label{tab:gr_prerl}
\begin{tabular}{lcccc}
\toprule
CPT & GSM8K & MATH-500 & GPQA & mean (8 tasks) \\
\midrule
none                         & 71.7 & 71.0 & 27.8 & 37.5 \\
OLMo2-1B-mined (transfer), 35M & 55.8 & 59.6 & 20.7 & 29.1 \\
targeted $n{=}3$, 35M        & 85.3 & 75.6 & 29.8 & 40.5 \\
targeted $n{=}3$, 100M       & 82.5 & 72.2 & 27.3 & 35.2 \\
targeted $n{=}3$, 1B         & 29.5 & 43.8 & 17.7 & 18.8 \\
targeted $n{=}5$, 35M        & 83.6 & 71.4 & 27.8 & 38.0 \\
targeted $n{=}5$, 100M       & 68.4 & 54.4 & 21.7 & 30.0 \\
targeted $n{=}5$, 1B         & 44.1 & 47.2 & 15.7 & 22.8 \\
random, 35M                  & 68.6 & 61.0 & 22.2 & 33.4 \\
random, 1B                   & 42.3 & 58.2 & 20.7 & 28.5 \\
mix, 500M $n{=}3$ $+$ 500M random & 20.6 & 43.4 & 13.6 & 18.8 \\
mix, 35M $n{=}3$ $+$ 965M random  & 47.1 & 61.6 & 11.6 & 27.4 \\
mix, 100M $n{=}3$ $+$ 900M random & 35.0 & 60.2 & 16.2 & 27.6 \\
\midrule
\multicolumn{5}{l}{\emph{Qwen3-1.7B-Base} (transfer $=$ the Qwen3-4B-mined $n{=}3$ corpus)}\\
none                              & 78.5 & 58.6 & 17.7 & 29.5 \\
transfer, 35M                     & 75.8 & 63.2 & 12.6 & 28.7 \\
transfer, 100M                    & 78.8 & 61.6 & 21.7 & \textbf{31.3} \\
transfer, 1B                      & 63.2 & 43.4 & 11.6 & 21.8 \\
self-mined $n{=}3$, 35M           & 71.6 & 65.2 & \phantom{0}7.6 & 28.7 \\
self-mined $n{=}3$, 100M          & 79.2 & 58.8 & 11.6 & 30.9 \\
self-mined $n{=}3$, 1B            & 65.9 & 42.8 & \phantom{0}5.6 & 21.5 \\
random, 1B                        & 34.4 & 47.2 & \phantom{0}0.0 & 18.0 \\
mix, 35M transfer $+$ 965M random & 56.9 & 58.2 & \phantom{0}0.5 & 26.0 \\
mix, 100M transfer $+$ 900M random& 62.9 & 58.8 & \phantom{0}0.5 & 26.3 \\
mix, 500M transfer $+$ 500M random& 64.7 & 54.0 & \phantom{0}4.5 & 25.8 \\
\bottomrule
\end{tabular}
\end{table}

\begin{table}[h]
\centering
\small
\caption{Replay sweep: uniformly sampled Olmo-mix documents mixed into the CPT corpus.
OLMo2-1B rows: final RL accuracy on GSM8K, mean $\pm$ s.d.\ over RL seeds ($n{=}3$ unless
marked; $^a$: two seeds, $59.7$--$59.9$; $^b$: single seed; $^c$: $34.3$ and $36.5$ on two
seeds, the third seed's RL never left $8\%$). Qwen3-4B row: the pre-RL eight-benchmark mean
of Table~\ref{tab:gr_prerl}. Sizes: after stage 1, random $=400$M, large $=435$M targeted;
released base and Qwen3-4B, random $=1$B, large $=500$M targeted $+$ $500$M random, and
``ours large'' $=1$B targeted. Multi-task row: macro average over the four tasks at RL step 595, mean $\pm$ s.d.\ over two RL seeds, self-mined $n{=}4$ corpus; the full targeted-fraction sweep is in Table~\ref{tab:replay_multi}.}
\label{tab:replay}
\resizebox{\linewidth}{!}{%
\begin{tabular}{lcccccc}
\toprule
setting & none & ours 35M & random only & 35M ours $+$ random & large ours $+$ random & large ours alone \\
\midrule
GSM8K / OLMo2-1B, after stage 1 & 37.4$\pm$1.1 & \textbf{38.2$\pm$1.5} & 35.5$\pm$1.0 & 35.4$^c$ & 35.5$\pm$0.6 & 33.4$^b$ \\
GSM8K / OLMo2-1B, released base & 59.3$\pm$0.7 & \textbf{60.6$\pm$0.9} & 58.5$^a$ & 59.9$^a$ & 59.7$^a$ & 59.9$^b$ \\
Qwen3-4B, before RL (8 tasks)   & 37.5 & \textbf{40.5} & 28.5 & 27.4 & 18.8 & 18.8 \\
multi-task / OLMo2-1B, released base & \textbf{52.6$\pm$0.4} & 51.5$\pm$0.1 & 51.4$\pm$1.0 & 51.4$\pm$0.3 & 51.9$\pm$1.3 & 51.7$\pm$0.4 \\
\bottomrule
\end{tabular}}
\end{table}

\begin{table}[h]
\centering
\small
\caption{Targeted fraction at a fixed $1$B-token CPT budget on the multi-task mixture
(OLMo2-1B released base, self-mined $n{=}4$ corpus, remainder uniformly sampled Olmo-mix
documents): final macro accuracy over the four tasks, mean $\pm$ s.d.\ over two RL seeds.
Reference points: no CPT $52.6\pm0.4$, the $35$M targeted slice alone $51.5\pm0.1$, matched
control $49.5$ (one seed).}
\label{tab:replay_multi}
\begin{tabular}{lccccccc}
\toprule
targeted tokens   & 0 & 35M & 100M & 200M & 500M & 700M & 1B \\
targeted fraction & 0\% & 3.5\% & 10\% & 20\% & 50\% & 70\% & 100\% \\
\midrule
accuracy & 51.4$\pm$1.0 & 51.4$\pm$0.3 & 52.1$\pm$0.6 & 51.8$\pm$0.2 & 51.9$\pm$1.3 & 51.9$\pm$0.6 & 51.7$\pm$0.4 \\
\bottomrule
\end{tabular}
\end{table}

\begin{table}[h]
\centering
\small
\caption{Alternative CPT data for the runs of Table~\ref{tab:targeting} (OLMo2-1B, GSM8K,
final RL accuracy; mean $\pm$ s.d.\ over three RL seeds where given). High-accuracy
$n$-grams: the retrieval pipeline run with the highest-$a(g)$ $n$-grams instead of the
failure set, i.e.\ targeting the states RL already handles; MegaMath-Web-Pro: $35$M tokens
of a curated math web corpus with no retrieval.}
\label{tab:targeting_rest}
\begin{tabular}{llcc}
\toprule
Insertion & CPT data & tokens & accuracy \\
\midrule
after stage 1 & high-accuracy $n$-grams   & 175M & 36.2 \\
after stage 1 & failure $n$-grams (ours)  & 175M & \textbf{37.7} \\
\midrule
released base & none                      & 0    & 59.3$\pm$0.7 \\
released base & high-accuracy $n$-grams   & 35M  & 59.4$\pm$1.2 \\
released base & MegaMath-Web-Pro (curated math) & 35M & \textbf{60.7} \\
released base & failure $n$-grams (ours)  & 35M  & \textbf{60.6$\pm$0.9} \\
\bottomrule
\end{tabular}
\end{table}

\section{State-level effect of CPT across settings}
\label{app:states}

Table~\ref{tab:states} measures, for every base/CPT pair we trained, how often the RL
policy enters the targeted failure states and how often it succeeds from them, before and
after the repair. $\mathcal{G}$ is the set of retrieval queries used to build that pair's
CPT corpus (10--17 $n$-grams), so the measurement asks exactly whether the states the
corpus targeted became more recoverable. Both members of a pair are compared at the same
RL step on the same prompts (the $G$ training rollouts of that step; for OLMo2-7B the
validation generations). Accuracy is the sampled training-rollout accuracy at that step, not
test accuracy.

\paragraph{Choice of RL steps.} The analysis sets of Table~\ref{tab:ngrams} and
Figure~\ref{fig:states} are mined in a window that starts after training-rollout accuracy
has plateaued (OLMo2-1B on GSM8K: $33\%$ at step 100 vs.\ $36\%$ at 210; Qwen3-4B: $25\%$
at 120 vs.\ $26\%$ at 226) and ends before the scoring window, which is the last 60 steps
shared by the compared runs: steps 120--140 and 150--210 (or 226) for the 210- and
226-step recipes, 480--520 and 535--595 for the 595-step multi-task recipe, and 50--60 and
80--104 for the Qwen3-1.7B runs, which stop at steps 103--105. The two windows are disjoint so that no $n$-gram is scored on the rollouts that
selected it (Section~\ref{sec:local}); they were fixed once per setting from the run length
and not tuned. Rows that compare runs of different lengths (OLMo2-7B, 226 vs.\ 208 steps)
use the last 27 steps of each run.

\begin{table}[h]
\centering
\small
\caption{Entry rate into the targeted failure states ($\Pr[E_{\mathcal{G}}]$, fraction of
rollouts containing any $g\in\mathcal{G}$), success rate from them ($a(\mathcal{G})$), and
mean $a(g)$ over $n$-grams with $\geq10$ occurrences in both runs ($k$ = number of such
$n$-grams), before $\to$ after CPT. $\dagger$: the Qwen3-1.7B multi-task set is dominated by a formatting fragment (\ngram{process
is that}, 664 $\to$ 3 occurrences). $\S$: $\mathcal{G}$ is the set of 23 failure
five-grams of Table~\ref{tab:ngrams}, mined from the no-CPT run at steps 480--520, rather
than the retrieval queries; the matched control enters them in $0.4\%$ of rollouts with
$6.2\%$ success.}
\label{tab:states}
\resizebox{\linewidth}{!}{%
\begin{tabular}{llrrrrr}
\toprule
Setting / model & CPT corpus & step & $\Pr[E_{\mathcal{G}}]$ & $a(\mathcal{G})$ & mean $a(g)$ ($k$) & rollout acc.\ \\
\midrule
GSM8K / OLMo2-1B, after stage 1 & failure $n$-grams, 35M & 150--210 & $5.7\to10.6$ & $25.7\to29.2$ & -- & $42.4\to43.7$ \\
GSM8K / OLMo2-1B, after stage 1 & matched control, 35M & 150--210 & $5.7\to6.7$ & $25.7\to28.7$ & -- & $42.4\to42.4$ \\
GSM8K / OLMo2-1B, after stage 1, RL seed 2 & failure $n$-grams, 35M & 150--210 & $4.7\to5.4$ & $23.0\to28.5$ & -- & $40.8\to44.5$ \\
GSM8K / OLMo2-1B, after stage 1, RL seed 3 & failure $n$-grams, 35M & 150--210 & $10.1\to6.9$ & $24.1\to27.4$ & -- & $43.6\to42.2$ \\
GSM8K / OLMo2-1B, released      & self-mined $n{=}5$, 35M & 210 & $1.1\to1.8$ & $18.5\to10.9$ & -- (1) & $78.3\to77.0$ \\
GSM8K / Qwen3-0.6B              & OLMo2-1B-mined, 35M    & 210 & $0.7\to1.1$ & $26.3\to22.2$ & -- (0) & $90.4\to91.7$ \\
GSM8K / Qwen3-4B                & OLMo2-1B-mined, 35M    & 210 & $0.5\to0.6$ & $100\to80$    & -- (1) & $98.2\to98.4$ \\
multi-task / OLMo2-1B           & self-mined $n{=}4$, 35M & 533 & $1.0\to1.2$ & $36.0\to15.6$ & -- (0) & $56.5\to54.5$ \\
multi-task / Qwen3-0.6B         & self-mined $n{=}4$, 35M & 595 & $1.8\to5.0$ & $59.6\to52.0$ & $92.0\to91.5$ (1) & $62.6\to63.0$ \\
multi-task / Qwen3-1.7B$^\dagger$ & self-mined $n{=}4$, 35M & 595 & $27.8\to10.6$ & $39.3\to84.6$ & $76.2\to61.8$ (1) & $69.8\to70.9$ \\
multi-task / Qwen3-4B           & OLMo2-1B-mined, 35M    & 595 & $1.5\to7.7$ & $57.9\to43.7$ & $59.5\to59.1$ (1) & $79.4\to80.2$ \\
multi-task / Qwen3-4B, mined five-grams$^\S$ & OLMo2-1B-mined, 35M & 535--595 & $2.0\to1.4$ & $10.8\to17.6$ & -- & $78.8\to79.1$ \\
multi-task / OLMo2-7B           & self-mined $n{=}6$, 35M & 208 & $2.2\to7.4$ & $64.0\to83.5$ & $51.9\to82.2$ (1) & $81.1\to81.5$ \\
General Reasoner / Qwen3-4B     & self-mined $n{=}3$, 35M & 66  & $7.3\to7.4$ & $41.2\to37.0$ & $31.0\to26.2$ (9) & $42.3\to41.8$ \\
General Reasoner / Qwen3-4B     & self-mined $n{=}5$, 35M & 67  & $6.5\to13.2$ & $36.7\to31.9$ & $17.8\to16.9$ (10) & $41.3\to37.7$ \\
General Reasoner / Qwen3-1.7B   & self-mined $n{=}3$, 35M & 103 & $25.6\to26.7$ & $26.6\to27.0$ & $25.6\to27.4$ (13) & $30.3\to29.9$ \\
General Reasoner / Qwen3-1.7B   & Qwen3-4B-mined $n{=}3$, 35M & 104 & $7.4\to7.3$ & $23.6\to27.3$ & $16.9\to27.7$ (9) & $31.8\to31.5$ \\
\bottomrule
\end{tabular}}
\end{table}

\begin{table}[h]
\centering
\small
\caption{Every retrieval query of the OLMo2-1B GSM8K corpus on all training rollouts of
steps 150--210 (stage-1 insertion, RL seed 1): number of rollouts containing the phrase
and their success rate, without CPT, with the matched control, and with our corpus. The
last two rows aggregate over the 16 queries and over the 66 five-grams of
Figure~\ref{fig:states}.}
\label{tab:queries_ag}
\begin{tabular}{lrrrrrr}
\toprule
 & \multicolumn{2}{c}{no CPT} & \multicolumn{2}{c}{matched control} & \multicolumn{2}{c}{ours} \\
query & $n$ & $a(g)$ & $n$ & $a(g)$ & $n$ & $a(g)$ \\
\midrule
\ngram{doesn't make sense}          & 1605 & 7.5\%  & 93  & 8.6\%  & 1486 & 5.4\% \\
\ngram{have a negative number of}   & 921  & 7.8\%  & 800 & 8.4\%  & 481  & 12.3\% \\
\ngram{there's a mistake}           & 745  & 6.2\%  & 20  & 15.0\% & 839  & 8.1\% \\
\ngram{be a mistake in}             & 731  & 8.3\%  & 46  & 6.5\%  & 684  & 6.1\% \\
\ngram{Let's correct}               & 706  & 5.2\%  & 123 & 1.6\%  & 289  & 4.5\% \\
\ngram{because we can't}            & 657  & 8.7\%  & 54  & 14.8\% & 55   & 1.8\% \\
\ngram{Let's recheck}               & 591  & 5.6\%  & 170 & 4.1\%  & 2166 & 6.3\% \\
\ngram{which is incorrect}          & 464  & 3.9\%  & 225 & 4.0\%  & 512  & 1.6\% \\
\ngram{However, based on the}       & 408  & 7.8\%  & 40  & 5.0\%  & 68   & 5.9\% \\
\ngram{we know it takes}            & 355  & 25.1\% & 337 & 45.1\% & 92   & 39.1\% \\
\ngram{need to recheck the}         & 320  & 5.9\%  & 91  & 3.3\%  & 402  & 4.7\% \\
\ngram{can't have half}             & 318  & 5.3\%  & 518 & 6.6\%  & 236  & 5.5\% \\
\ngram{we'll round}                 & 208  & 7.2\%  & 43  & 0.0\%  & 54   & 1.9\% \\
\ngram{there's a misunderstanding}  & 160  & 6.9\%  & 6   & 0.0\%  & 158  & 8.9\% \\
\ngram{or a mistake in}             & 158  & 8.9\%  & 6   & 0.0\%  & 76   & 14.5\% \\
\ngram{However, let's}              & 144  & 6.2\%  & 3   & 0.0\%  & 11   & 9.1\% \\
\midrule
any of the 16 queries           & 3063 & 9.6\%  & 2054 & 13.1\% & 3247 & 8.6\% \\
any of the 66 mined five-grams  & 8859 & 25.7\% & 10498 & 28.7\% & 16620 & 29.2\% \\
\bottomrule
\end{tabular}
\end{table}

Across the three stage-1 RL seeds, the success rate from the failure states rises on every
seed ($+3.5$, $+5.5$, $+3.3$) while the entry rate rises on seeds 1 and 2 and falls on
seed 3, whose final test accuracy is also the one below no CPT ($36.5$ vs.\ $37.9$).

Three patterns stand out. (i) Only the before-mid-training setting shows a rise in
$a(\mathcal{G})$ together with an accuracy gain, and there the failure set is mined on
steps 120--140 and evaluated on 150--210 to avoid the selection bias that a
final-step comparison carries; the rise is modest ($+6$ points, control $+2.5$) and comes
with a doubled entry rate. (ii) On released base models the targeted
states are rare ($\leq2\%$ of rollouts) and the CPT corpus tends to make the model emit the
trigger phrases \emph{more} often (e.g., Qwen3-4B multi-task: \ngram{still incorrect} 0 $\to$
162 occurrences at 36\% accuracy), which raises the entry rate without raising the success
rate. (iii) In the General Reasoner setting the self-mined operation $n$-grams remain
low-value after CPT, whereas the transferred corpus raises their value without changing
accuracy, i.e., recoverability of a state and task accuracy can move independently.

\end{document}